\pdfoutput=1

\documentclass[11pt]{article}

\usepackage[preprint]{acl}

\usepackage{times}
\usepackage{latexsym}

\usepackage[T1]{fontenc}

\usepackage[utf8]{inputenc}

\usepackage{microtype}

\usepackage{inconsolata}

\usepackage{amsmath,amsfonts,bm}

\def\eqref#1{equation~\ref{#1}}

\def\1{\bm{1}}

\DeclareMathAlphabet{\mathsfit}{\encodingdefault}{\sfdefault}{m}{sl}
\SetMathAlphabet{\mathsfit}{bold}{\encodingdefault}{\sfdefault}{bx}{n}

\usepackage{subcaption}
\usepackage{fancyhdr}
\usepackage{graphicx}
\usepackage{tikz}
\usepackage{listings}
\usepackage{color}
\usepackage{booktabs}
\usepackage{multirow}
\usepackage{color, colortbl}
\usepackage[linesnumbered,ruled,vlined]{algorithm2e}
\usepackage{comment}
\usepackage{amsmath,amssymb} 
\usepackage{xcolor}     
\usepackage{color, colortbl}
\definecolor{Gray}{gray}{0.9}
\hypersetup{
    colorlinks,
    linkcolor={blue!50!black},
    citecolor={blue!96!red!62!green!96!},
    urlcolor={blue!80!black}
}
\usepackage{epsfig}
\usepackage{subcaption}
\usepackage{boldline}
\def\*#1{\mathbf{#1}}
\usepackage{amssymb}
\usepackage{pifont}
\newcommand{\cmark}{\ding{51}}%
\newcommand{\xmark}{\ding{55}}%
\usepackage{hyperref}
\usepackage{url}
\usepackage{xcolor}     
\usepackage[linesnumbered,ruled,vlined]{algorithm2e}
\usepackage{algpseudocode}
\usepackage{algorithmicx}

\usepackage{color, colortbl}
\definecolor{Gray}{gray}{0.9}
\definecolor{myblue}{HTML}{d8ebf8}

\def\name{\textbf{$\texttt{Int}$}CoOp\xspace}
\definecolor{COLOR_ZS}{HTML}{E6ECE3}

\title{\texttt{Int}CoOp: Interpretability-Aware Vision-Language Prompt Tuning}

\author{Soumya Suvra Ghosal \and Samyadeep Basu \and Soheil Feizi \and Dinesh Manocha\\
         University of Maryland, College Park \\
         \{\texttt{sghosal, sbasu, sfeizi, dmanocha}\}@umd.edu}

\begin{document}
\maketitle
\begin{abstract}

Image-text contrastive models such as CLIP learn transferable and robust representations for zero-shot transfer to a variety of downstream tasks. However, to obtain strong downstream performances, prompts need to be carefully curated, which can be a tedious engineering task. To address the issue of manual prompt engineering, prompt-tuning is used where a set of contextual vectors are learned by leveraging information from the training data. Despite their effectiveness, existing prompt-tuning frameworks often lack interpretability, thus limiting their ability to understand the compositional nature of images. In this work, we first identify that incorporating compositional attributes (e.g., a {\it ``green''} tree frog) in the design of manual prompts can significantly enhance image-text alignment scores. Building upon this observation, we propose a novel and interpretable prompt-tuning method named~\name, which learns to jointly align attribute-level inductive biases and class embeddings during prompt-tuning. To assess the effectiveness of our approach, we evaluate \name across two representative tasks in a few-shot learning setup: generalization to novel classes, and unseen domain shifts. Through extensive experiments across 10 downstream datasets on CLIP, we find that introducing attribute-level inductive biases leads to superior performance against state-of-art prompt tuning frameworks. Notably, in a 16-shot setup, \name improves CoOp by $7.35\%$ in average performance across $10$ diverse datasets. 
\end{abstract}

\section{Introduction}

\label{sec:intro}

Recently, significant advancements have been achieved in the field of vision-language models, with notable examples like CLIP~\citep{radford2021learning}, Flamingo~\citep{alayrac2022flamingo}, ALIGN~\citep{jia2021scaling}, and CoCa~\citep{yu2022coca}. These models have excelled in acquiring transferable and robust image representations, a feat accomplished through a combination of two fundamental components:  (i) Large-scale paired image-text datasets ranging from 400M to 2B image-text pairs; (ii) A contrastive objective that aligns the image and text embeddings into a common subspace. Leveraging these ingredients, vision-language models have obtained strong performances in zero-shot classification, image-text retrieval, and robustness to distribution shifts. For all these tasks, contrastive models such as CLIP enable zero-shot inference: Given an image $\mathcal{I}$ and a set of text prompts $\{t_{i}\}_{i=1}^{N}$, the most relevant text-prompt $t\in\{t_{i}\}_{i=1}^{N}$ is identified by maximizing the image-text similarity between $\mathcal{I}$ and $t$. 

Adapting image-text contrastive models for downstream tasks is a complex undertaking. Achieving optimal performance with image-text contrastive models necessitates the manual creation of domain-specific prompts, a process that demands extensive domain knowledge and is exceptionally challenging and time-consuming. Even with considerable prompt engineering, there is limited assurance that the designed prompt is truly optimal. To address this issue, recent research~\citep{zhou2022conditional, lee2023read, khattak2023maple, ouali2023black} has turned to prompt-tuning techniques, borrowing concepts from the field of NLP and applying them to vision-language models like CLIP to achieve good image recognition performance on downstream tasks. However these frameworks often \emph{lack interpretability} and as a result the model struggles to understand the composition of the images.

\begin{figure*}[t]
    \centering
    \includegraphics[width = \textwidth]{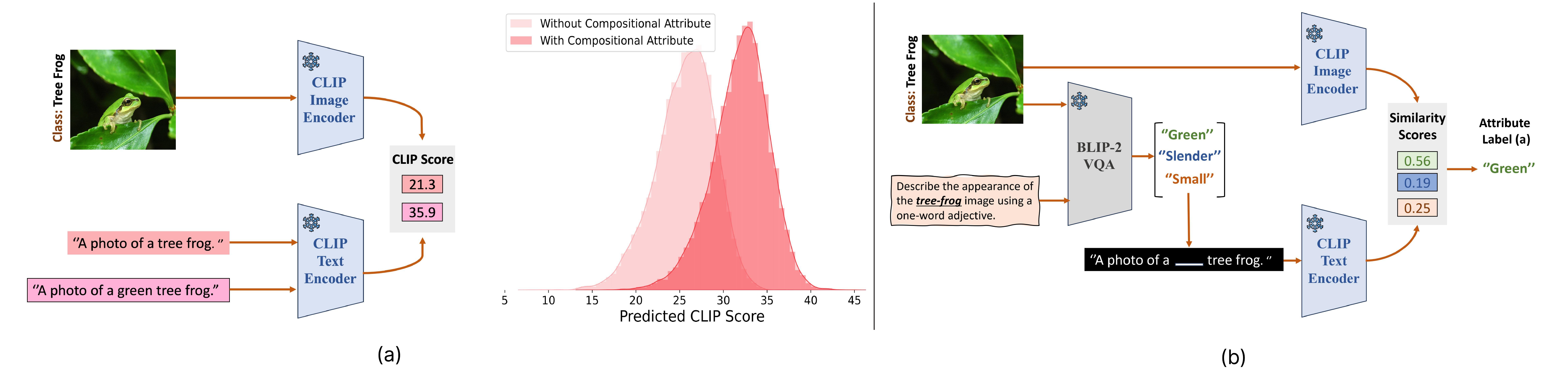}
    \caption{\small \textbf{(a) Importance of learning interpretable concepts in prompts.}  \textbf{Left:} For each image, we design two prompt templates: (1) Without any compositional attribute ``A photo of a $[cls]$'' and (2) With compositional information ``A photo of a $[a]$ $[cls]$'' where $[cls]$ represents the classname and $[a]$ represents an attribute obtained using a BLIP-2 based VQA model. \textbf{Right:} The distribution plot highlights the importance of baking attribute information into the prompts. For this analysis, we used a CLIP model with a ViT-B/16 image encoder and a dataset consisting of $50$ images selected randomly from each of $1000$ classes in ImageNet-1k. The x-axis indicates the predicted CLIP score. Clearly, the CLIP model is more confident when the prompts include information related to the compositionality of the image. \textbf{(b) Framework for obtaining attribute-level supervision}. We present the overarching architecture for generating attribute labels $a$ for a given training image using BLIP-2 VQA model.}

    \label{fig:conf_plot}
    \vspace{-0.5cm}
\end{figure*}
In this study, we address this challenge by learning a method to extract and embed attribute-level information into the prompt-tuning framework during training. We define an \emph{attribute} as an interpretable concept that is relevant to the image and encapsulates its semantic essence. Although manually crafted prompts can vary in their characteristics based on the specific downstream domain, our analysis has revealed a noteworthy trend. We have observed that prompts containing attribute information that describes the objects in the images lead to enhanced image-text alignment scores in contrastive models such as CLIP. For instance, as depicted in Figure~\ref{fig:conf_plot}, we can see that prompts incorporating compositional attributes such as ``green'' tree frog yield higher image-text alignment scores than those lacking such descriptors.

Based on these findings, we present an interpretable prompt-tuning approach known as \name, which incorporates attribute information into the prompt-tuning procedure thereby generating more interpretable prompts. While one might initially consider leveraging off-the-shelf image captioning models to generate attribute labels, this approach becomes infeasible during inference when class labels are unavailable. Consequently, generating attribute descriptions for images emerges as a \emph{non-trivial task}. To mitigate this challenge, we train a compact hypernetwork responsible for predicting embeddings corresponding to attribute descriptors.

We test our prompt-tuning method~\name on a range of diverse downstream datasets to test for generalization to novel classes, and robustness to distribution shifts.  In Section~\ref{sec:results}, we show that our method~\name has improved robustness to distribution shifts, domain generalization, and few-shot learning. Notably, in domain generalization setup, \name outperforms PLOT~\cite{chen2023plot} by $19.32\%$ in average performance across 4 diverse domains. In summary, our research provides compelling empirical support for the substantial advantages of integrating attribute-level inductive biases into the prompt-tuning process.

Overall, our paper makes the following key contributions:
\begin{itemize}
\vspace{-0.2cm}
    \item We introduce a novel prompt-tuning method, named~\name, which concurrently aligns attribute-level inductive biases and class embeddings during training, thus facilitating the generation of interpretable prompts. 
    \vspace{-0.2cm}
    \item We devise an efficient cross-attention mechanism to integrate image information with the learnable prompt tokens seamlessly.
    \vspace{-0.2cm}
     \item We present comprehensive experiments across a range of tasks, including generalization to unseen classes, and distribution shifts showing the efficacy of~\name. Notably, in a $16-$shot setup, \name outperforms the state-of-art framework LFA~\cite{ouali2023black} by $1.27\%$ improvement in average performance across $10$ diverse datasets.
   
\end{itemize}

\section{Related Works}
\label{sec:related_works}
\textbf{Pretrained Vision-Language Models.}
Recent research~\citep{radford2021learning, yu2022coca} has shown that leveraging language to train image encoders can result in strong downstream performances especially for robustness and few-shot learning. These vision-language models are usually pre-trained on large corpuses of image-text pairs using contrastive objectives that align image and text representations into a common subspace. CLIP~\citep{radford2021learning} and ALIGN~\citep{align} use {\it only} the contrastive objective to align image-text embeddings. CoCa~\citep{yu2022coca} uses a captioning loss in conjunction with contrastive objectives to further improve image representations. For example, CLIP is pre-trained on $\sim$400M image-text pairs whereas ALIGN is pre-trained on a much larger set of $\sim$1B image-text pairs. In recent times, masked vision-language objectives~\citep{kwon2023masked} have also resulted in strong image-text representations.

However, in all these vision-language models, inference requires manually curated prompts to extract the best performance, which can be a tedious engineering task. To mitigate this issue, recent research has turned to prompt-tuning techniques to automatically learn domain specific prompts.

\vspace{0.1cm}
\noindent\textbf{Prompt Tuning.} 
Given a set of text-instructions and an image, existing vision-language models make their decisions by selecting the text-instruction which has the maximum similarity between the image and text-embeddings. 

Recent advances in this field, such as methods like CoOp~\citep{zhou2022learning}, CoCoOp~\citep{zhou2022conditional}, ProDA~\citep{lu2022prompt}, VPT~\citep{jia2022visual}, MaPLe~\cite{khattak2023maple}, KgCoOp~\citep{yao2023visual}, ProGrad~\citep{zhu2022prompt}, LASP~\cite{bulat2023lasp}, RPO~\cite{lee2023read}, DAPT~\cite{cho2023distribution}, PLOT~\cite{chen2023plot}, and LFA~\cite{ouali2023black} have shifted from manually designed prompts to automatically learning prompts through fine-tuning learnable vectors with image-text pairs from the target domain. CoOp fine-tunes CLIP to optimize a set of learnable tokens in the input layer of the text-encoder. CoCoOp enhances CoOp by incorporating conditional image information in the prompt-learning process. VPT learns tokens in each layer of a given encoder through a fine-tuning objective. KgCoOp introduces a regularizer to constrain the prompt tuning process  such that the representations of the learned prompts do not deviate significantly from the manually crafted prompts. ProGrad leverages the prompt gradients to fine-tune the learnable tokens such that the prior knowledge in the vision-language model is preserved. PLOT applies optimal transport to
match the vision and text modalities for generating the
discriminative and visual-aligned local textual prompt tokens. Refer ~\citet{liu2024few} for a comprehensive survey on prompt-tuning frameworks. Overall, none of the existing works aim to understand if augmenting certain inductive biases in the prompt-tuning process is beneficial. Our work~\name specifically addresses this issue and shows that incorporating compositional attributes in the prompt-tuning process can indeed be beneficial for downstream tasks.

\section{Preliminaries}
\label{subsec:prelims}

\paragraph{Contrastive Language-Image Pre-training (CLIP)~\citep{radford2021learning}} is a vision-language model trained on a large dataset of $400$ million image-text caption pairs using a contrastive loss. CLIP primarily consists of two major components: 

\noindent(1) \textbf{Vision Encoder $\mathcal{V}(\cdot)$} consists of a ViT~\citep{vit} model, which takes an image $\mathcal{I} \in \mathbb{R}^{H \times W \times 3}$ as input and outputs a visual embedding in the latent space. The vision encoder $\mathcal{V}$ consists of $L$ transformer blocks $\left\{\mathcal{V}_{i}\right\}_{i=1}^{L}$. First, the input image $\mathcal{I}$ is split into $R$ fixed-size patches which are projected into patch embeddings $E_{0} \in \mathbb{R}^{R \times D_v}$, where $D_v$ is the constant latent vector size of the image encoder. Patch embeddings $E_{i}$ are input to the $(i+1)^{\text{th }}$ transformer block $\left(\mathcal{V}_{i+1}\right)$ along with a learnable class token $\mathbf{x}_{i}$
and is sequentially processed through $L$ transformer blocks:
\begin{equation*}
    \left[\mathbf{x}_{i}, E_{i}\right]=\mathcal{V}_{i}\left(\left[\mathbf{x}_{i-1}, E_{i-1}\right]\right) \quad i=1,2, \cdots, L .
\end{equation*}

To obtain the final image representation, the class token $\mathbf{x}_{L}$ of the last transformer layer $\left(\mathcal{V}_{L}\right)$ is projected to a common image-text latent embedding space via a linear projection layer.

\begin{equation*}
    \mathcal{V}(\mathcal{I}) =\texttt{Proj}\left(\mathbf{x}_{L}\right) \quad \mathbf{x}_{L} \in \mathbb{R}^{D_{vl}} .
\end{equation*}

where ${D_{vl}}$ is the constant vector size of the  image-text latent embedding space.

\noindent(2) \textbf{Text Encoder $\mathcal{T}(\cdot)$} is a transformer-based model that maps the input text captions into text embeddings. 

For zero-shot inference on a downstream dataset consisting of $C$ classes with class names $\{[cls]_c\}_{c=1}^C$, CLIP uses hand-crafted prompts to generate the textual class embeddings. Specifically, given a hand-crafted prompt template ``A photo of a $[cls]$'', let $\*s_c$ represent the sequence embedding for the prompt ``A photo of a $[cls]_c$'' corresponding to the $c$-th class. Given an input image $\mathcal{I}$, the output probability is given by:
\begin{equation}
    \mathbb{P}(\hat{y}=c|\mathcal{I}) = \frac{\text{exp}(\text{cos}(\mathcal{V}(\mathcal{I}), \mathcal{T}(\*s_c))/\tau)}{\sum_{j = 1}^C\text{exp}(\text{cos}(\mathcal{V}(\mathcal{I}), \mathcal{T}(\*s_j))/\tau)}
\end{equation}
where $\text{cos}(\cdot,\cdot)$ represents the cosine similarity and $\tau$ is the temperature coefficient.

\paragraph{Context Optimization (CoOp)~\citep{zhou2022learning}.} Designing hand-crafted prompts in CLIP for every downstream data set is a tedious and time-consuming task. To mitigate this issue of prompt engineering, CoOp~\citep{zhou2022learning} proposed to learn the prompts directly from the data by replacing the hand-crafted prompt with a context vector comprising of $M$ tunable vectors. Let the context vector be represented as $\*u = \{\*u_1, \*u_2, \cdots, \*u_M\}$, where $\*u_i$ represents a $512$-dimensional vector\footnote{The vector $\*u_i$ is of same dimension as the word-embedding of class names $[cls]_c$. In this study, we primarily use CLIP-ViTB/16 model where text embeddings are projected in a $512$-dimensional space.}. Unlike the hand-crafted prompt template, the tunable prompts are now designed as $\*p = \{[\*u_1, \*u_2, \cdots, \*u_M, [cls]_c]\}_{c=1}^C$. To allow the exchange of information learned from the data, the context vector $\*u$ is common across all the class categories. Finally, the context vector $\*u$ is learned by minimizing the cross-entropy loss between the ground-truth and predicted label as follows:
\vspace{-0.5cm}
\begin{align}
\small
     \mathbb{P}(\hat{y}=c|\mathcal{I}) &= \frac{\text{exp}(\text{cos}(\mathcal{V}(\mathcal{I}), \mathcal{T}(\*p_c))/\tau)}{\sum_{j = 1}^C\text{exp}(\text{cos}(\mathcal{V}(\mathcal{I}), \mathcal{T}(\*p_j))/\tau)} \\
    \mathcal{L}_{\text{CE}} &= -\text{log}~\mathbb{P}(\hat{y} = y|\mathcal{I})
\end{align}
where, $y$ represents the true label for image $\mathcal{I}$ and $\*p_c$ represents the tunable prompt for class $c$. Note that during training \name, the vision and text encoder in CLIP are completely \emph{frozen} and the optimization framework only updates the context vector $\*u$.

\section{\texttt{Int}CoOp: Interpretability-Aware Prompt Tuning}
\label{subsec:comp}

\begin{figure}[t]
    \centering
    \includegraphics[width = \columnwidth]{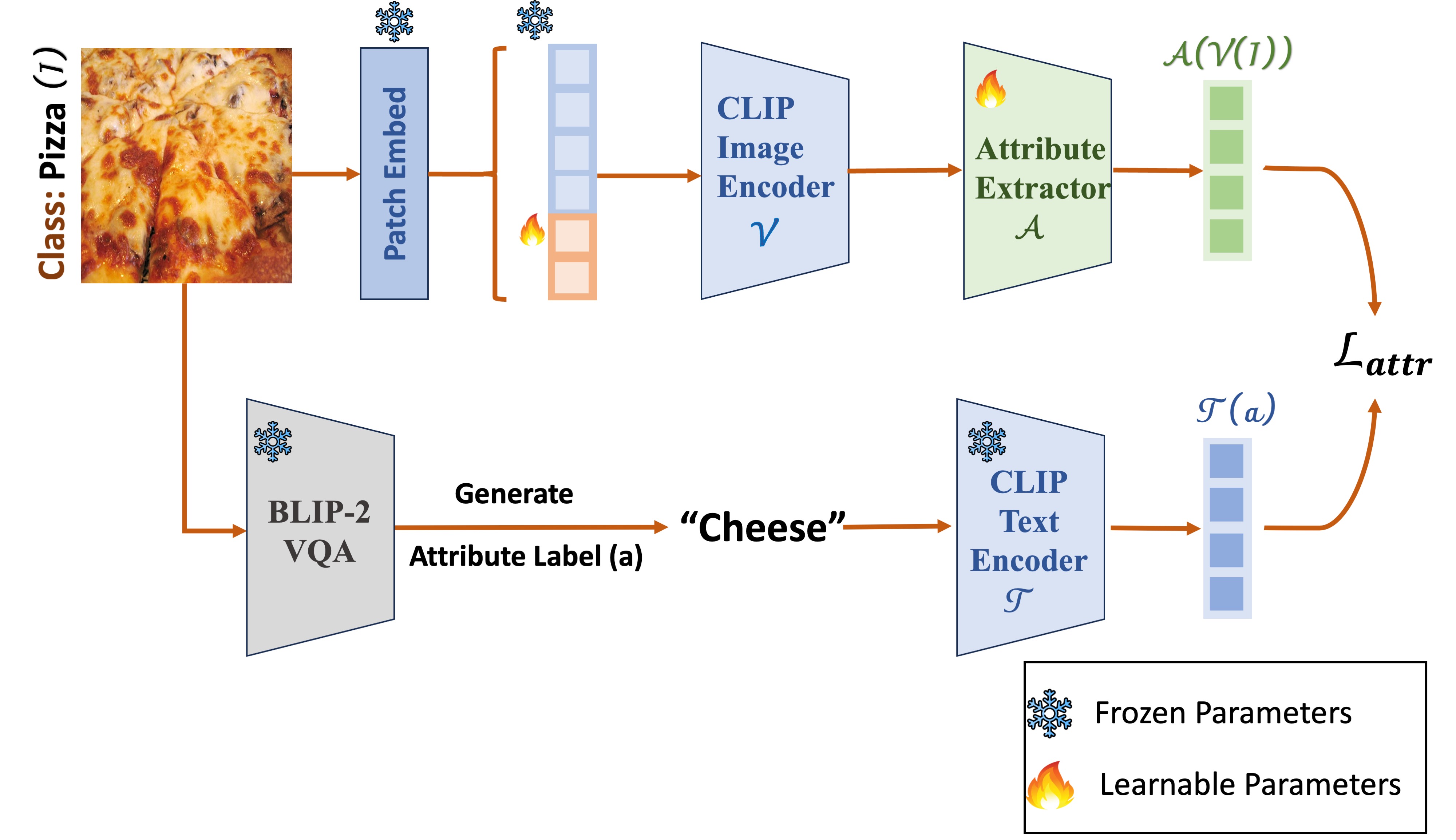}
    \caption{\small \textbf{Framework for learning compositional attributes.} The figure elucidates the training framework of the attribute extractor network $\mathcal{A}$. }
    \label{fig:attribute_framework}
\end{figure}

In this section, we provide a detailed overview of our proposed prompt-tuning approach \name. In Section~\ref{subsec:learn_prompts}, we detail the process of extracting attribute information from a given image. Next, in Section~\ref{subsec:cond_prompts}, we delve deeper to understand the process of generating image-conditioned prompts. Finally, we outline our entire training framework in Section~\ref{subsec:training}, demonstrating the integration of all components into the training pipeline. Similar to past context optimization approaches~\cite{zhou2022learning}, \name can also be easily applied to a broad family of CLIP-like vision-language models.

\subsection{Learning Interpretable Image Concepts}
\label{subsec:learn_prompts}

\vspace{0.2cm}
\noindent\textbf{Obtaining Attribute-level Supervision.} Given an input image $\mathcal{I}$, our goal is to extract an interpretable attribute (denoted by $a$) that provides an accurate characterization of the image. For example, given the image of ``Tree Frog'' in Figure~\ref{fig:conf_plot}(b), we can define the attribute $a$ as ``Green''. However, standard image-recognition datasets such as Imagenet~\citep{deng2009imagenet} only provide true labels for object categories and do not consist of attribute-level supervision. We overcome this problem by using a BLIP-2~\citep{li2023blip2} ViT-G FlanT5XXL based VQA model to generate an attribute label ($a_{\mathcal{I}}$) for each image $\mathcal{I}$ in the train set. The entire framework is visually represented in Figure~\ref{fig:conf_plot}(b). We refer the reader to the Appendix~\ref{app:blip_extension} for a detailed description and visualization of more representative examples.

\vspace{0.2cm}
\noindent\textbf{Learning to extract attribute information during training.} During inference, as class labels are unavailable for test images, direct utilization of off-the-shelf captioning models~\citep{li2023blip2} is infeasible. To circumvent this limitation, we propose training a network to learn contextually relevant attributes (see Figure~\ref{fig:attribute_framework}). Specifically, we design an attribute extractor network $\mathcal{A}$, which takes as input the image embedding from CLIP's vision encoder and outputs a $512$-dimensional vector representing the embedding of the attribute. This network is trained using supervised attribute labels obtained from the framework in Figure~\ref{fig:conf_plot}(b).

\vspace{0.2cm}
\noindent\textbf{Designing the attribute extractor.} It is important to note that the attribute extractor network $\mathcal{A}$ learns the interpretable concepts directly from the image embedding. Therefore, the embedding vector must effectively encode information regarding the compositionality of the image to enable proper training of the network. In Table~\ref{tab:ablation_1}, we show that the embeddings from CLIP's frozen vision encoder are not expressive enough to essentially capture the attribute information. This challenge is compounded by the fact that, in a few-shot setup, there are a limited number of samples available for each class, leading to suboptimal training of the attribute extractor. To generate richer and more informative visual representations, we append a set of $n$ learnable parameters $\{\*Z^{j}_i \in \mathbb{R}^{D_v}\}_{j=1}^n$ to each transformer layer $\mathcal{V}_i$ of the image encoder up to depth $K$~\citep{jia2022visual, khattak2023maple}.
\vspace{-0.2cm}
\begin{multline*}
    \left[\mathbf{x}_{i}, E_{i}, \textunderscore \right]=\mathcal{V}_{i}\left(\left[\mathbf{x}_{i-1}, E_{i-1}, Z_{i-1}\right]\right)\\
    \forall \small{i=1,2, \cdots, K.}
\end{multline*}
\vspace{-1cm}
\begin{multline*}
    \left[\mathbf{x}_{j}, E_{j}, Z_{j} \right]=\mathcal{V}_{i}\left(\left[\mathbf{x}_{j-1}, E_{j-1}, Z_{j-1}\right]\right)\\
    \forall {j=K+1, \cdots, L.}
\end{multline*}
\begin{equation*}
    \mathcal{V}(\mathcal{I}) =\texttt{Proj}\left(\mathbf{x}_{L}\right)
\end{equation*}

In Section~\ref{sec:ablation}, we show that this improved design choice leads to better performance on downstream tasks. Finally, the generated attribute labels can be used to train the network $\mathcal{A}$ by minimizing the following loss:
\begin{equation}
    \mathcal{L}_{\text{attr}} = || \mathcal{A}(\mathcal{V}(\mathcal{I})) -  \mathcal{T}(a_{\mathcal{I}})||_f^f
\end{equation}
where $||\cdot||_f^f$ indicates the $f$-th norm, $ \mathcal{T}(a_{\mathcal{I}})$ represents the $512$-dimensional token embedding of the attribute $a_{\mathcal{I}}$. In Appendix~\ref{app:ablation_design}, based on ablations we find setting $f=2$ gives the best performance. In this paper, we instantiate the network $\mathcal{A}$ with a two-layer neural net with ReLU activations.

\subsection{Instance-Conditional Prompts}
\label{subsec:cond_prompts}

In this section, we delve deeper into understanding how the prompts are generated. Recall from Section~\ref{subsec:prelims}, that for CoOp~\cite{zhou2022learning}, the context vector $\*u = \{\*u_1, \*u_2, \cdots, \*u_M\}$ is shared across all classes, and the tunable prompts are designed as $\*p = \{[\*u_1, \*u_2, \cdots, \*u_M, [cls]_c]\}_{c=1}^C$. In Table~\ref{tab:ablation_1}, we show that sharing the context vectors across all images leads to sub-optimal generalization to novel classes. To address this concern, we opt for a strategy that involves generating instance-conditional context tokens. However, rather than a straightforward addition of the image embedding to the context tokens~\citep{zhou2022conditional}, we employ a Multi-head Attention module. This module generates context tokens by attending to the image embedding. Given an input image $\mathcal{I}$, the image attended context vector $\*h(\mathcal{I})$ is given by:

\begin{multline*}
     \*h(\mathcal{I}) = \texttt{MultiHead}(\text{Query=}\*u, \text{Key=}\mathcal{V}(\mathcal{I}), \\ \text{Value=}\mathcal{V}(\mathcal{I}))
\end{multline*}

\noindent where $\*u$ represents the context vector, and \texttt{MultiHead} indicates a Multi-head attention module. Note that the instance-conditioned context vector $\*h(\mathcal{I})$ has the same shape as $\*u$.

Finally, we can generate the prompts for each class by embedding the output of the attribute extractor into the instance-conditioned context vector $\*h(\mathcal{I})$. Let ${\*p}^{+}(\mathcal{I})$ represent the attribute incorporated prompts and is defined as:

\vspace{-0.2cm}
\begin{equation}
\label{eq:prompt_gen}
    {\*p}^{+}(\mathcal{I}) = \{[\*h_1, \cdots, \*h_M, \mathcal{A}({\mathcal{V}}(\mathcal{I})), [cls]_c]\}_{c=1}^C
\end{equation}
Unlike prior works~\citep{zhou2022conditional}, our cross-attention based image-conditioning mechanism incorporates a learned weighted sum of various points in the image embedding for a single position in the context vector, thereby providing a stronger conditioning signal. In Section~\ref{sec:ablation}, we empirically show that our conditioning mechanism is better suited for few-shot fine-tuning in CLIP.

\subsection{Regularizing the Prompts}
\label{subsec:regularization}

Analysis by \citet{yao2023visual} reveal that without any regularization, the context vectors may heavily overfit the training data. This can lead to poor performance on unseen classes during inference. To mitigate this, they propose adding a knowledge-guided loss that aims to minimize the discrepancy between the learned prompts and the handcrafted template ``A photo of a $[cls]$''. In this paper, we also add an additional loss term to regularize the learned prompts. However, instead of simply using the hand-crafted templates, we generate a set of textual prompts incorporating the compositional information for each image. Given an image $\mathcal{I}$, let $\{\*p^{\text{gen}}_i(\mathcal{I})\}_{i=1}^N$ represent the pool of $N$ synthetically generated prompt templates embedded with interpretable concepts $a_{\mathcal{I}}$ in image $\mathcal{I}$. In this study, we select $N=80$ diverse textual prompts as suggested in \citet{radford2021learning}. Based on this, we define the regularization loss as:
\begin{equation}
    \mathcal{L}_{\text{reg}} = \frac{1}{N} \sum_{i=1}^N ||\mathcal{T}({\*p}^{+}(\mathcal{I})_y) - \mathcal{T}(\*p^{\text{gen}}_i(\mathcal{I}))||_g^g
\end{equation}
where $y$ represents the true label for the image $\mathcal{I}$, $\mathcal{T}(\cdot)$ is the CLIP text encoder and ${\*p}^{+}(\mathcal{I})_y = [\*h_1, \cdots, \*h_M, \mathcal{A}(\mathcal{V}(\mathcal{I})), [cls]_y]$ is the learnable prompt for the true class $y$. Based on ablations in Appendix~\ref{app:ablation_design}, we set $g=1$. 

\subsection{Putting it together}
\label{subsec:training}

Let $\mathcal{D}^{\text{train}}=\{\mathcal{I}_j,y_j\}_{j=1}^J$ represent a training dataset consisting of $J$ samples, where $\mathcal{I}_j$ is an image and $y_j \in \{1,\cdots,C\}$ represents the corresponding label. Given the dataset, we first generate the attribute labels ($a_{\mathcal{I}}$) for each image as defined in Section~\ref{subsec:learn_prompts}. Note, to avoid any computational overhead during training, we perform this operation offline. Based on the previous discussions, the training loss is formulated as:
\begin{equation}
    \mathcal{L} = \mathcal{L}_{\text{CE}} + \lambda_1\mathcal{L}_{\text{attr}} + \lambda_2\mathcal{L}_{\text{reg}} 
\end{equation}  
\vspace{-1.1cm}
 \begin{multline*}
  \small
     \text{where}~\mathcal{L}_{\text{CE}}= \\ \small -\frac{1}{J}\sum_{j=1}^J \text{log}\frac{\text{exp}(\text{cos}(\mathcal{V}(\mathcal{I}_j), \mathcal{T}({\*p}^{+}(\mathcal{I}_j)_{y_j}))/\tau)}{\sum_{c = 1}^C\text{exp}(\text{cos}(\mathcal{V}(\mathcal{I}_j), \mathcal{T}({\*p}^{+}(\mathcal{I}_j)_c))/\tau)} 
 \end{multline*}

where $y_j$ represents the true label for the image $\mathcal{I}_j$ and $C$ represents the number of seen classes. The optimization framework aims to learn the optimal parameters by minimizing the training loss as 
$\min~\mathbb{E}_{(\mathcal{I},y)\sim \mathcal{D}^{\text{train}}}~[\mathcal{L}]$. Based on ablations in Appendix~\ref{app:ablation_design}, we set $\lambda_1 = 4$ and $\lambda_2 = 4$.

\section{Experiments}
\label{sec:results}

\noindent\textbf{Implementation Details:} In this study, for all experimentation, we use a pretrained CLIP~\cite{radford2021learning} model with a ViT-B/16 image encoder unless otherwise specified. We train the model for $50$ epochs using a batch size of $4$ and SGD optimizer with a learning rate of $0.0025$. We set the context length $M=4$. Further, for training \name, we append $n=4$ learnable visual tokens in each transformer layer upto a depth of $K=9$. We report results averaged over $3$ random seeds. All experiments are run using the configurations listed in Appendix~\ref{app:hardware}. The code will be made publicly available following paper acceptance.

\vspace{0.3cm}
\noindent\textbf{Computational Efficiency:} In Table~\ref{tabapp:comp_eff} (Appendix), we compare the computational cost of training and inference for \name compared to baseline framework such as CoOp~\citep{zhou2022learning}. We observe that, due to instance-conditional prompt generation, \name's per-epoch training time is slightly higher compared to CoOp. However, we believe this minor increase in training time is justified by the significant performance improvements shown in Table~\ref{tab:main_table}. During inference, as presented in Table~\ref{tabapp:comp_eff}, \name does not incur any significant additional overhead compared to CoOp.

\subsection{Base-to-Novel Class Generalization} 
\label{subsec:base_novel}

Following existing literature~\cite{zhou2022learning, zhou2022conditional, yao2023visual}, to assess the generalization capability of \name, we employ a zero-shot setting that involves partitioning datasets into base and novel classes. Our model is exclusively trained on the base classes within a few-shot framework, and its performance is evaluated across both the base and novel categories.

\vspace{0.3cm}
\noindent\textbf{Datasets:} To evaluate on generalization from base-to-novel classes, in line with past studies~\cite{zhou2022learning}, we used 10 diverse image classification datasets: ImageNet~\cite{deng2009imagenet},  Caltech101~\cite{fei2004learning}, OxfordPets~\cite{parkhi2012cats}, StanfordCars~\cite{cars}, Flowers102~\cite{nilsback2008automated}, Food101~\cite{bossard2014food},  FGVCAircraft~\cite{maji2013fine}, SUN397~\cite{xiao2010sun}, UCF101~\cite{soomro2012ucf101}, and EuroSAT~\cite{helber2019eurosat}. We refer the reader to Table~\ref{tab:description} (Appendix) for a detailed description of the datasets used in this study.

\newcolumntype{?}{!{\vrule width 1pt}}
\newcolumntype{a}{>{\columncolor{myblue}}c}
\begin{table*}[t]
      \centering
       
        \resizebox{\textwidth}{!}{%
        \begin{tabular}{lcccccccccccc?a}
        \toprule
       Dataset & Set & CLIP & CoOp & Co-CoOp  & MaPLe & KgCoOp & ProGrad & LASP & RPO  & DAPT & PLOT & LFA & \name  \\

       & & & \small (IJCV22) & \small (CVPR22) & \small (CVPR23) & \small (CVPR23) & \small (ICCV23) & \small (ICCV23) & \small 
 \small (ICCV23) & \small (ICCV23) & \small (ICLR23) &  \small (ICCV23) & (Ours) \\

\midrule
       \multirow{3}{*}{ImageNet} & Base & 72.43 & 76.47 & 75.98 & 76.66 & 75.83 & 77.02 & 76.20 & 76.60 & 76.83 & 77.30 & 76.89 & 75.99 \\
       & Novel & 68.14 & 67.88 & 70.43 & 70.54 & 69.96 & 66.66 & 70.95 & 71.57 & 69.27 & 69.87 & 69.36 & 72.67 \\
       
     & HM &  70.22 & 71.92 & 73.10 & 73.47 & 72.78 & 71.46 & 73.48 & 74.00 & 72.85 & 73.40 & 72.93 & \textbf{74.29} \\
      \midrule 
      \multirow{3}{*}{Caltech101} & Base & 96.84 & 98.00 & 97.96 & 97.74 & 97.72 & 98.02 & 98.10 & 96.03 & 97.83 & 98.53 & 98.41 & 97.80 \\
      & Novel & 94.00 & 89.91 & 93.81 & 94.36 & 94.39 & 93.89 & 94.24 & 94.37 & 93.81 & 92.80 & 93.93 & 94.76\\
      & HM & 95.40 & 93.73 & 95.84 & 96.02 & 96.03 & 95.91 & 96.16 & 96.03 & 95.39 & 95.58 & 96.13 & \textbf{96.25} \\
      \midrule
       \multirow{3}{*}{OxfordPets} & Base & 91.17 & 93.67 & 95.20 & 95.43 & 94.65 & 95.07 & 95.90 & 94.63 & 95.00 & 94.50 & 95.13 & 95.92 \\
     &  Novel & 97.26 & 95.29 & 97.69 & 97.76 & 94.65 & 95.07 & 97.93 & 97.50 & 95.83 & 96.83 & 96.23 & 98.20 \\
     &  HM & 94.12 & 94.47 & 96.43 & 96.58 & 96.18 & 96.33 & 96.90 & 96.05 & 95.41 & 95.65 & 95.68 & \textbf{97.04} \\
      \midrule
     \multirow{3}{*}{Stanford Cars} & Base & 63.37 & 78.12 & 70.49 & 72.94 & 71.76 & 77.68 & 75.17 & 74.69 & 75.80 & 78.57 & 76.32 & 
 77.04 \\
     & Novel & 74.89 & 60.40 & 73.59 & 74.00 & 75.04 & 68.63 & 71.60 & 75.53 & 63.93 & 74.80 & 74.88 & 76.32 \\
     & HM & 68.65 & 68.13 & 72.01 & 73.47 & 73.36 & 72.88 & 73.34 & 74.69 & 69.36 & 76.63 & 75.59 & \textbf{76.67} \\
     \midrule
     \multirow{3}{*}{Flowers102} & Base & 72.08 & 97.60 & 94.87 & 95.92 & 95.00 & 95.54 & 97.00 & 94.13 & 96.97 & 97.93 & 97.34 & 97.82\\
     & Novel & 77.80 & 59.67 & 71.75 & 72.46 & 74.73 & 71.87 & 73.53 & 76.67 & 60.90 & 74.00 & 75.44 & 75.54 \\
     & HM & 74.83 & 74.06 & 81.71 & 82.56 & 83.65 & 82.03 & 83.95 & 84.50 & 74.81 & 83.99 & 85.00 & \textbf{85.24} \\
     \midrule
     \multirow{3}{*}{Food101} & Base & 90.10 & 88.33 & 90.70 & 90.71 & 90.50 & 90.37 & 91.20 & 90.33 & 90.37 & 89.80 & 90.52 & 91.45 \\
     & Novel & 91.22 & 82.26 & 91.29 & 92.05 & 91.70 & 89.59 & 91.70 & 90.33 & 91.30 & 91.37 & 91.48 & 91.99 \\
     & HM & 90.66 & 85.19 & 90.99 & 91.38 & 91.09 & 89.98 & 91.44 & 90.58 & 90.83 & 90.58 & 91.00 & \textbf{91.72}\\
     \midrule
     \multirow{3}{*}{FGVC Aircraft} & Base & 27.19 & 40.44 & 33.41 & 37.44 & 36.21 & 40.54 & 34.53 & 37.33 & 39.97 & 42.13 & 41.48 & 38.55 \\
     & Novel & 36.29 & 22.30 & 23.71 & 35.61 & 33.55 & 27.57 & 30.57 & 34.20 & 29.80 & 33.73 & 32.29 & 35.90  \\
     & HM & 31.09 & 28.75 & 27.74 & 36.50 & 34.83 & 32.82 & 32.43 & 35.70 & 34.14 & \textbf{37.46} & 36.31 & 37.17 \\
     \midrule
     \multirow{3}{*}{SUN397} & Base & 69.36 & 80.60 & 79.74 & 79.75 &  80.29 & 81.26 & 80.70 & 80.60 & 78.92 & 77.68 & 79.59 & 81.63 \\
     & Novel & 75.35 & 65.89 & 76.86 & 78.70 & 76.53 & 74.17 & 78.60 & 77.80 & 76.97 & 73.63 & 77.20 & 79.33\\
     & HM & 72.23 & 72.51 & 78.27 & 79.75 & 78.36 & 77.55 & 79.63 & 79.18 & 78.92 & 77.68 & 79.59 & \textbf{80.46} \\
     \midrule
  
     \multirow{3}{*}{EuroSAT} & Base & 56.48 & 92.19 & 87.49 & 94.07 & 85.64 & 90.11 & 94.60 & 86.63 & 94.73 & 93.70 & 93.40 & 95.26\\
     & Novel & 64.05 & 54.74 & 60.04 & 73.23 & 64.34 & 60.89 & 77.78 & 76.79 & 50.33 & 62.67 & 71.24 & 78.01 \\
     & HM & 60.03 & 68.69 & 71.21 & 82.30 & 73.48 & 72.67 & 85.36 & 76.79 & 65.74 & 75.11 & 80.83 & \textbf{85.77} \\
     \midrule
     \multirow{3}{*}{UCF101} & Base & 70.53 & 84.69 & 82.33 & 83.00  & 82.89 & 84.33 & 84.77 & 83.67 & 84.30 & 86.60 & 86.97 & 86.76 \\
     & Novel & 77.50 & 56.05 & 73.45 & 78.66 & 76.67 & 74.94 & 78.03 & 79.34 & 76.33 & 75.90 & 77.48 & 79.42 \\
     & HM & 73.85 & 67.46 & 77.64 & 80.77 & 79.65 & 79.35 & 81.26 & 79.34 & 80.12 & 80.90 & 81.95 & \textbf{82.92} \\
    \midrule      
    Average & HM & 73.23 & 73.40 & 77.98 & 79.28 & 78.27 & 77.53 & 79.35 & 78.69 & 75.75 & 78.69 & 79.48 & \textbf{80.75} \\
      
      \bottomrule
    \end{tabular}%
        }
 \caption{\small \textbf{Comparison with state-of-art on base-to-novel generalization}. We observe that \name consistently demonstrates superior performance over existing prompt-tuning methods. HM represents the harmonic mean of the base and novel accuracies. We train all methods with $16$-shots samples from the base classes.}
    \label{tab:main_table}
\end{table*}

\vspace{0.3cm}
\noindent\textbf{\name outperforms the state-of-art.} In Table~\ref{tab:main_table}, we compare the base-to-new generalization ability of \name with baselines such as zero-shot CLIP and competitive prompt tuning frameworks such as CoOp~\citep{zhou2022learning}, CoCoOp~\citep{zhou2022conditional}, MaPLe~\cite{khattak2023maple}, KgCoOp~\citep{yao2023visual}, ProGrad~\citep{zhu2022prompt}, LASP~\cite{bulat2023lasp}, RPO~\cite{lee2023read}, DAPT~\cite{cho2023distribution}, PLOT~\cite{chen2023plot}, and LFA~\cite{ouali2023black} on a set of $10$ diverse datasets. We implemented all methods using a few-shot training approach involving $16$ randomly sampled shots for each base class. Recall that for this task, evaluation involves training the model solely on the base classes and assessing its performance on both base and novel classes, a challenging scenario that tests the model's generalizability. We employ the harmonic mean (HM) of the base and novel accuracies as the metric for comparison. Our empirical findings reveal two key insights: (1) \name consistently demonstrates superior few-shot performance in comparison to the state-of-the-art prompt tuning techniques. Moreover, when considering the average mean performance across all 10 datasets, \name outperforms the current state-of-art~\citep{ouali2023black} by $1.27\%$. Further, it also surpasses CoOp~\cite{jia2022visual}, a baseline prompt tuning framework, by $7.52\%$. (2) \name's strong performance is particularly evident in datasets featuring images with well-defined attributes, such as ImageNet, Flowers102, OxfordPets, StanfordCars and Caltech-101. For instance, on the OxfordPets dataset, \name enhances the novel accuracy by $1.97\%$ and $3.55\%$ compared to LFA and KgCoOp respectively. 

\subsection{Domain Generalization}
\label{subsec:domain}

To evaluate domain generalization, we utilized ImageNet~\cite{deng2009imagenet} as the source dataset and four of its variants as target datasets. These variants included ImageNetV2~\cite{recht2019imagenet}, ImageNetSketch~\cite{wang2019learning}, ImageNet-A~\cite{hendrycks2021natural}, and ImageNet-R~\cite{hendrycks2021many}, contributing to a comprehensive examination of domain shift scenarios. Our findings in Table~\ref{tab:domain_gen} indicate that \name demonstrates superior performance across all target datasets. Notably, \name improves the average accuracy by $1.41\%$ and $19.32\%$ compared to ProGrad and PLOT respectively. These results underscore the significance of learning interpretable attributes within the prompts. 

In Table~\ref{tab:few-shot} (Appendix), we also evaluate the generalizability of our proposed method on a 4-shot setting. Across all datasets considered, \name outperforms all compared methods on average. Overall, we find that~\name leads to strong and improved performances on a range of downstream tasks including novel class generalization, robustness to distribution shifts and few-shot learning, while being more interpretable than other prompt-tuning methods. 

\newcolumntype{?}{!{\vrule width 1pt}}
\begin{table}[!t]
\centering
\resizebox{\columnwidth}{!}{%
\begin{tabular}{lc?cccc?c}
\toprule
 & \textbf{Source} & \multicolumn{4}{c}{\textbf{Target}} \\
\midrule
 & ImageNet & -V2 & -Sketch & -A & -R & Avg. \\
\midrule
CLIP & 66.73 & 60.83 & 46.15 & 47.77 & 73.96 & 57.18 \\
CoOp & 71.51 & 64.20 & 47.99 & 49.71 & 75.21 & 59.27\\
CoCoOp & 71.02 & 64.07 & 48.75 & 50.63 & 76.18 & 59.90 \\
MaPLe & 70.72 & 64.07 & 49.15 &  50.90 & 76.98 & 60.28 \\
KgCoOp & 71.20 & 64.10 &  48.97 & 50.69 & 76.70 & 60.11\\
ProGrad & 72.24 & 64.73 & 47.61 & 49.39 & 74.58 & 59.08\\
LASP & 71.10 & 63.96 & 49.01 & 50.70 & 77.07 & 60.19 \\
RPO & 71.76 & 65.13 & 49.27 & 50.13 & 76.57 & 60.27 \\
DAPT &  72.20 & 64.93 & 48.30 & 48.74 & 75.75 & 59.43 \\ 
PLOT & 63.01 & 55.11 & 33.00 & 21.86 & 55.61 & 41.39 \\
LFA & 72.65 & 64.72 & 48.01 & 51.50 & 76.09 & 60.08 \\
\midrule
\rowcolor{myblue}\name (Ours) & 71.85 & 65.21 & 49.20 & 51.55 & 76.88 & \textbf{60.71}  \\
\bottomrule
\end{tabular}%
}
\caption{\small\textbf{\name leads to improved performances on domain generalization tasks.} 
The model is trained on ImageNet~\cite{deng2009imagenet} dataset in a few-shot setup with 16 samples per class and evaluated on four domain-shifted ImageNet datasets.}
\label{tab:domain_gen}
\end{table}

\section{Discussion}
\label{sec:discussion}

\begin{figure}[!t]
    \centering
    \includegraphics[width=\linewidth]{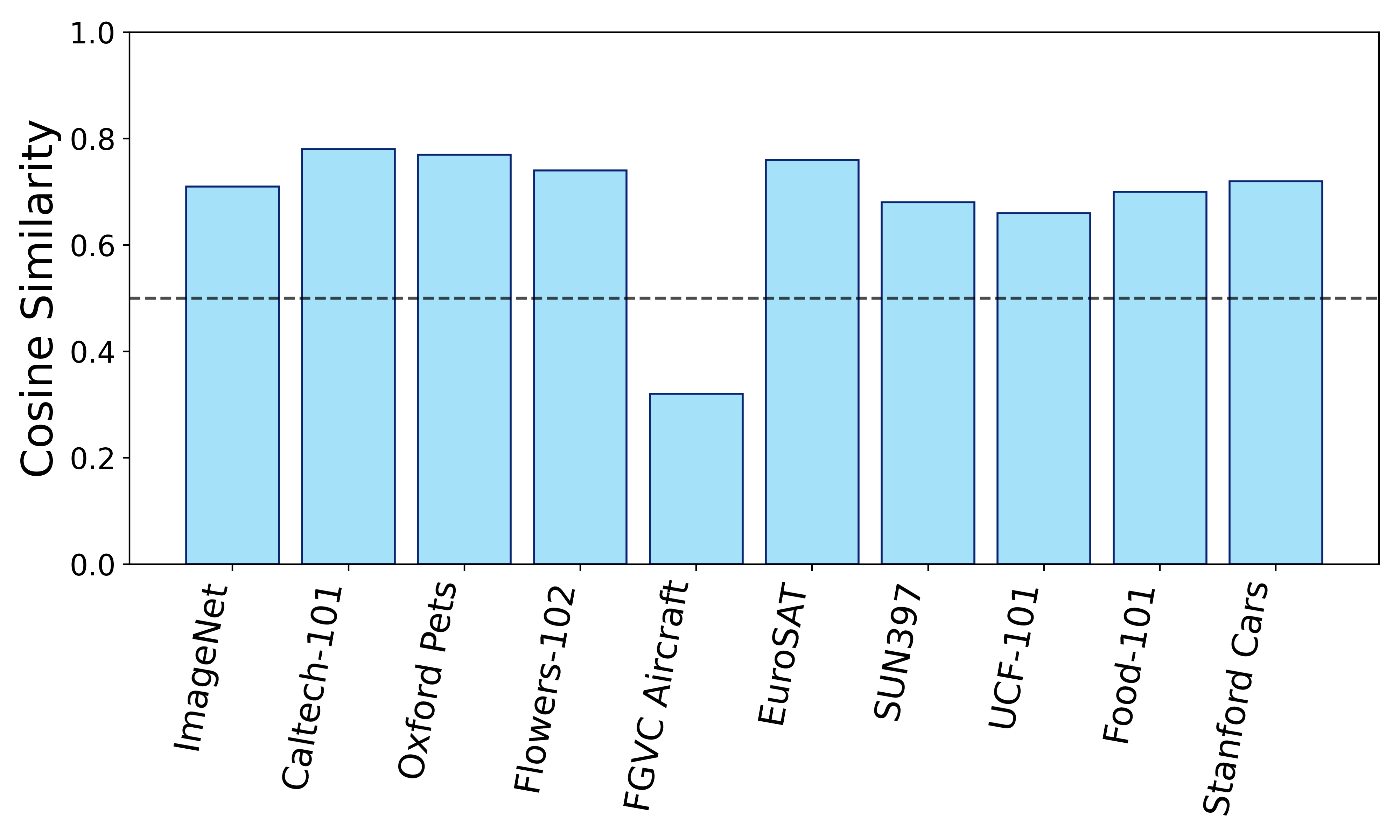}
    \caption{We measure the cosine similarity between the learned attribute embedding $\mathcal{A}(\mathcal{V}(\mathcal{I}))$ and the BLIP-2 generated label $a_{\mathcal{I}}$. A high cosine similarity indicates that \name effectively learns contextually relevant attributes.}
    \label{fig:cosine_sim_direct}
\end{figure}

\paragraph{\name learns interpretable prompts.} In this section, we delve deeper into understanding the quality of the attributes generated by \name during inference. Given a test image $\mathcal{I}$ with true label $y$, we first extract its corresponding learned attribute embedding $\mathcal{A}(\mathcal{V}(\mathcal{I}))$. To evaluate the quality of this embedding, we utilize the BLIP-2 model to produce an attribute label $a_{\mathcal{I}}$. We evaluate two setups: (1) Firstly, to validate the quality of the attributes generated by \name, in Figure~\ref{fig:cosine_sim_direct}, we visualize the cosine similarity of the learned attribute embedding $\mathcal{A}(\mathcal{V}(\mathcal{I}))$ and the BLIP-2 generated label $a_{\mathcal{I}}$. Across all datasets, we observe a high similarity between the generated attribute embedding and the BLIP-2-generated label. This confirms that \name effectively learns contextually relevant and correct attribute information. (2) Secondly, as illustrated in Figure~\ref{fig:analyse_prompt} (Appendix), we observe that the prompts crafted using the learned attribute embedding $\mathcal{A}(\mathcal{V}(\mathcal{I}))$ closely align with the original prompt format ``A photo of $[a]$ $[cls]$'', as evidenced by high cosine similarity. On the other side, prompts lacking the attribute information exhibit reduced similarity. This analysis highlights that during inference, \name generates prompts with interpretable compositional information, thereby explaining the improved performance.

\vspace{0.1cm}
\noindent\textbf{Importance of learning meaningful attributes.} In this section, we further validate the importance of learning contextually meaningful attributes during training. To illustrate this, we experiment by substituting the original attribute labels generated by the BLIP-2 model for each image in the training set with irrelevant adjectives. Specifically, we exchange the attribute labels among different classes, ensuring each image is paired with an unrelated adjective through careful human supervision. For instance, in the altered setup, the image labeled as a ``cheese pizza'' in Figure~\ref{fig:attribute_framework} is mislabeled as a ``green pizza'', where the attribute ``green'' bears no relevance to the image. Employing the experimental framework as described in Section~\ref{subsec:base_novel}, this alteration results in an HM accuracy of 63.27\% on the ImageNet-1k dataset— a decline of 11.02\% compared to the performance achieved with \name. This significant drop in accuracy highlights the critical role of learning accurate and relevant attributes in training.

\noindent\emph{For additional discussion, we refer the reader to Appendix~\ref{app:discussion}.}

\section{Ablations on Design Choice}
\label{sec:ablation}

In this section, we delve into a comprehensive exploration of the design choices made in our proposed framework. 

\vspace{0.1cm}
\noindent\textbf{Ablations on Visual Prompting.} As illustrated in Section~\ref{subsec:learn_prompts}, to enhance image representations \name effectively utilizes the deep visual prompting approach. To substantiate our design rationale, we conduct ablation experiments as outlined in Table~\ref{tab:ablation_1} (Appendix). From our empirical analysis, we make two key observations: (1) Visual prompting plays a crucial role in training \name. Specifically, training without any visual prompting, where the frozen CLIP embeddings are used to train the attribute network $\mathcal{A}$, leads to notably inferior performance. (2) Appending visual tokens to deeper transformer layers provides a substantial performance boost in average performance compared to a shallow prompting strategy. 

\vspace{0.1cm}
\noindent\textbf{Ablations on Instance Conditioning.} To condition the prompts on the input image, prior studies~\cite{zhou2022conditional} have proposed the direct addition of the image embedding to the context vector. However, as elaborated in Section~\ref{subsec:cond_prompts}, we employ a multi-head attention module for generating image-conditioned prompts in the training of \name. In Table~\ref{tab:ablation_1} (Appendix), we present empirical results that bolster the importance of utilizing an attention-based conditioning approach in contrast to additive conditioning. Specifically, we observe a $1.58\%$ improvement in average performance when using a Multihead attention based conditioning.

\section{Conclusion}
\label{sec:conclusion}

In our paper, we initially observe that incorporating relevant attributes into prompts significantly improves image-text alignment in CLIP. To achieve this enhancement, we present a novel technique called \name, which integrates these attributes into learned prompts. This integration is made possible by leveraging a BLIP-2~\citep{li2023blip2} model to annotate attributes in few-shot datasets.
With the image as a conditioning factor, we devise a hypernetwork responsible for predicting embeddings corresponding to attribute descriptors. Simultaneously, we optimize the other context vectors using CLIP's contrastive objective. 
 Our comprehensive testing across diverse datasets underscores the significant improvement in zero-shot performance achieved by \name.

\section{Limitations}
\label{app:limitations}

Our study, through its extensive evaluation across multiple datasets, demonstrates that augmenting prompts with attribute information can substantially enhance CLIP's effectiveness in various downstream applications. However, our approach has certain limitations: (1) A notable constraint of our approach is that its effectiveness may diminish in scenarios where images are devoid of specific attribute-level details. Despite this, it is noteworthy that in practical, real-world contexts, such as with the ImageNet dataset, \name consistently outperforms its counterparts. (2) The performance of \name is contingent upon the quality of attributes generated for images in the training set. Poorly generated attributes can detrimentally affect performance. 

For future work, we plan to investigate improved attribute extraction techniques to handle images with less discernible attribute-level details and to generate attributes with greater diversity.

\bibliography{main}

\clearpage
\newpage
\appendix

\section{Software and Hardware}
\label{app:hardware}
We run all experiments with Python 3.7.4 and PyTorch 1.9.0. For all experimentation, we use two Nvidia RTX 2080-Ti and a single A5000 GPU.

\section{Extension: Obtaining Attribute-level Supervision}
\label{app:blip_extension}
In Section 3.2.1 of the main paper, we demonstrated how the generated attribute labels can be used for training \name. In this section, we will provide a more detailed explanation of the procedure for extracting attribute labels for an image. In this paper, we leverage a BLIP-2 ViT-G FlanT5XXL visual question-answering (VQA) model for zero-shot generation of attribute labels. Specifically, given an image $\mathcal{I}$ with class label $[cls]$, we employ the templates shown in Table~\ref{tab:template} to prompt the VQA model to generate $3$ captions corresponding to each image. To improve caption variety, we generate these captions under varying random seeds and set \texttt{repetition\_penalty}$=100$ to discourage repetitive outputs. Note that the prompt templates for each dataset have been manually tuned with some domain information to improve performance. Subsequently, we select the most suitable caption based on the CLIP score. In Figure~\ref{fig:rep_image_1} and Figure~\ref{fig:rep_image_2}, we show some representative images from various datasets and the corresponding generated attributes.

\section{Note on Attributes Generated by BLIP-2}
To understand the effectiveness of BLIP-2 in correctly annotating few-shot tasks with their adjectives - we designed a proxy task with $215$ images, where each image is labeled with its attribute. Given that it is difficult to perform a scalable manual annotation of attributes, we take advantage of first pre-defining captions which contain an adjective describing an object, and then generating corresponding images from them. The object list is a subset from MS-COCO -- namely $O = $\{\texttt{handbag}, \texttt{pizza}, \texttt{suitcase}, \texttt{bottle}, \texttt{firehydrant}, \texttt{cup}, \texttt{cake}, \texttt{book}, \texttt{vase}, \texttt{cat} \}. The attribute list for each object $o \in O$ is created by prompting ChatGPT with prompts such as:  {\it 'Describe some of the possible \textbf{shapes} of object o in one word', 'Describe some of the possible \textbf{colors} of object o in one word'....}. These attributes from ChatGPT are then filtered and quality-controlled by our team to make sure that the attributes from ChatGPT are relevant to the object $o \in O$. 
Leveraging prompts in the template of ``A photo of a $[a]$ $[o]$'', we then generate 215 images from \texttt{Stable-Diffusion-v2}~\cite{Rombach_2022_CVPR} in total across all the classes, where $[a]$ represents the attribute label and $[o]$ is the object name.
Across these generated images, we then prompt BLIP-2 with prompts such as: {\it 'Describe the \textbf{shape} of the object in one word', 'Describe the \textbf{color} of the object in one word' ....} to predict the attribute. Subsequently, we measured the cosine similarity between BLIP-2's predictions and the ground truth attribute labels $a$. Given that there are only $215$ images in our validation set, in addition to the qualitative analysis, we also manually compared the BLIP-2 predicted attributes and the ground truth to check the effectiveness of BLIP-2. Our investigation revealed a compelling 85\% similarity between BLIP-2 predictions and the ground truth. This highlights that BLIP-2 is a suitable candidate to generate attributes for annotation of few-shot datasets. 

\newcolumntype{?}{!{\vrule width 1pt}}
\newcolumntype{a}{>{\columncolor{myblue}}c}
\begin{table}[!t]
\centering
\resizebox{0.7\columnwidth}{!}{%
\begin{tabular}{lc?a}
\hline
Datasets & Oracle & \name\\
\hline
ImageNet & 74.37 & 74.29 \\
Caltech101 & 96.00 &  96.25\\
OxfordPets & 97.13 & 97.04 \\
StanfordCars & 76.67 & 76.67 \\
Flowers102 & 85.32 & 85.24  \\
Food101 & 91.66 & 91.72\\
FGVCAircraft & 36.99 & 37.17 \\
SUN397 & 80.50 & 80.46 \\
EuroSAT & 85.80 & 85.77\\
UCF101 &  82.96 & 82.92 \\
\hline
Avg. & 80.74 & 80.75 \\
\hline
\end{tabular}%
}
\caption{Comparing \name's average performance with oracle setup as described in Appendix~\ref{app:discussion} across 10 datasets.}
\label{tab:oracle_setup}
\end{table}
\begin{table*}[!t]
      \centering
      \resizebox{0.8\textwidth}{!}{%
        \begin{tabular}{lccc}
        \toprule
        \textbf{Methods} & Train Time (in mins) & Inference Time (in mins) & HM  \\
        \midrule
        CoOp~\citep{zhou2022learning} & 1.03 & 0.032 & 94.47 \\
        \rowcolor{myblue}\name & 2.15 & 0.041 & \textbf{97.04} \textbf{\textcolor{red}{(+2.57)}} \\
        \bottomrule
        \end{tabular}}
        \caption{\textbf{Computational Efficiency of \name.} We compare the training and inference time of \name with CoOp~\citep{zhou2022learning}. For training time, we report the duration taken to train for one epoch on the Oxford Pets dataset~\citep{parkhi2012cats}. Similarly, for inference time, we report the duration taken to infer on a test image from the Oxford Pets dataset. The numbers reported are averaged for 3 different runs.}
    \label{tabapp:comp_eff}
\end{table*}
\newcolumntype{?}{!{\vrule width 1pt}}
\begin{table*}[t]
    \centering
    \resizebox{\textwidth}{!}{%
    \begin{tabular}{lc}
    \toprule
    \textbf{Dataset} & \textbf{Prompt Template} \\
    \midrule
    ImageNet & ``Describe the appearance of the $[cls]$ image using a one-word adjective.'' \\
    Caltech-101 & ``Describe the appearance of the $[cls]$ image using a one-word adjective.'' \\
    OxfordPets & ``Describe a one-word adjective such as color for the $[cls]$ image''. \\
    Flowers102 &  ``Describe the color of the $[cls]$ flower in one word.'' \\
    FGVCAircraft & ``Describe a one-word adjective for the aircraft image.''\\
    StanfordCars & ``Describe a one-word adjective for the $[cls]$ car image.'' \\
    Food101 &  ``Describe a one-word adjective for the $[cls]$ food image.''    \\
    SUN397 &  "Describe a one-word adjective summarizing the appearance of the $[cls]$ image.''\\
    EuroSAT &  ``Describe a one-word adjective that best describes the natural surroundings in this satellite image of $[cls]$.'' \\
    UCF101 & ``Describe a one-word adjective describing the action of the person in this $[cls]$ image.'' \\
    \bottomrule
    \end{tabular}%
    }
    \caption{Templates used for prompting the BLIP-2 model for different datasets. $[cls]$ represents the class name for the given image.}
    \label{tab:template}
\end{table*}
\section{Extension: Results on Few-shot Learning}

To further evaluate the generalizability of our proposed method, we conducted experiments on a 4-shot setting. In this case, the model is trained on only $4$ samples from each base class. We report the average accuracy over base and novel classes in Table~\ref{tab:few-shot}. We observe that under a $4$-shot setup, \name consistently outperforms state-of-art prompt tuning approaches across multiple datasets. Notably, on OxfordPets, \name enhances the average performance by $3.45\%$ and $3.83\%$ compared to PLOT~\cite{chen2023plot} and DAPT~\cite{cho2023distribution}. Across all datasets considered, \name outperforms all compared methods on average.

\section{Extension: Additional Discussion }
\label{app:discussion}

To further understand the efficiency of the attribute extractor, we compare \name's performance with the following setup: we directly use the BLIP-2 embedding $\mathcal{T}(a_{\mathcal{I}})$ in Equation~\ref{eq:prompt_gen} to train our framework, keeping all other losses the same. Specifically, during training, the BLIP-2 generated attribute embeddings are directly integrated into the prompts instead of using the output from the attribute extractor $\mathcal{A}$. However, during inference, since the class labels are unavailable, we utilize the trained attribute extractor to generate descriptions for test images. We refer to this setup as the \emph{oracle} setting, as it uses the true labels during training. The results for this setup are reported in Table~\ref{tab:oracle_setup}. Notably, the performance obtained using the oracle setting is almost identical to \name's performance. This indicates that using the true attribute labels during training provides no additional advantage. Therefore, we can conclude that during training, the attribute extractor network $\mathcal{A}$ successfully learns to mimic the BLIP-2 embeddings, thereby generating interpretable prompts.

\newcolumntype{?}{!{\vrule width 1pt}}
\begin{table}[t]
    \centering
    \resizebox{\columnwidth}{!}{%
    \begin{tabular}{lcc?cc?c}
    \toprule
    & \multicolumn{2}{c?}{Visual Prompting} & \multicolumn{2}{c?}{Instance Conditioning} & \multirow{2}{*}{HM} \\
    \cmidrule{2-5}
    & Shallow (K=1) & Deep (K=9)  & Additive~\cite{zhou2022conditional} & Multihead &  \\
    \midrule
    & \cmark & \xmark & \xmark & \xmark & 75.01  \\
    & \xmark & \cmark & \xmark & \xmark & 76.90 \\
    \midrule
    & \xmark & \xmark & \cmark & \xmark &  74.31 \\
    & \xmark & \xmark & \xmark & \cmark &   75.89\\
    \midrule
\rowcolor{myblue} \name (Ours)    & \xmark & \cmark & \xmark & \cmark & \textbf{80.75}\\
    \end{tabular}%
    }
    \caption{\small\textbf{Ablation on design choices.} We perform ablation experiments to delineate the importance of each component in our proposed approach.}
    \label{tab:ablation_1}
\end{table}

\begin{figure}[t]
    \centering
    \includegraphics[width=\columnwidth]{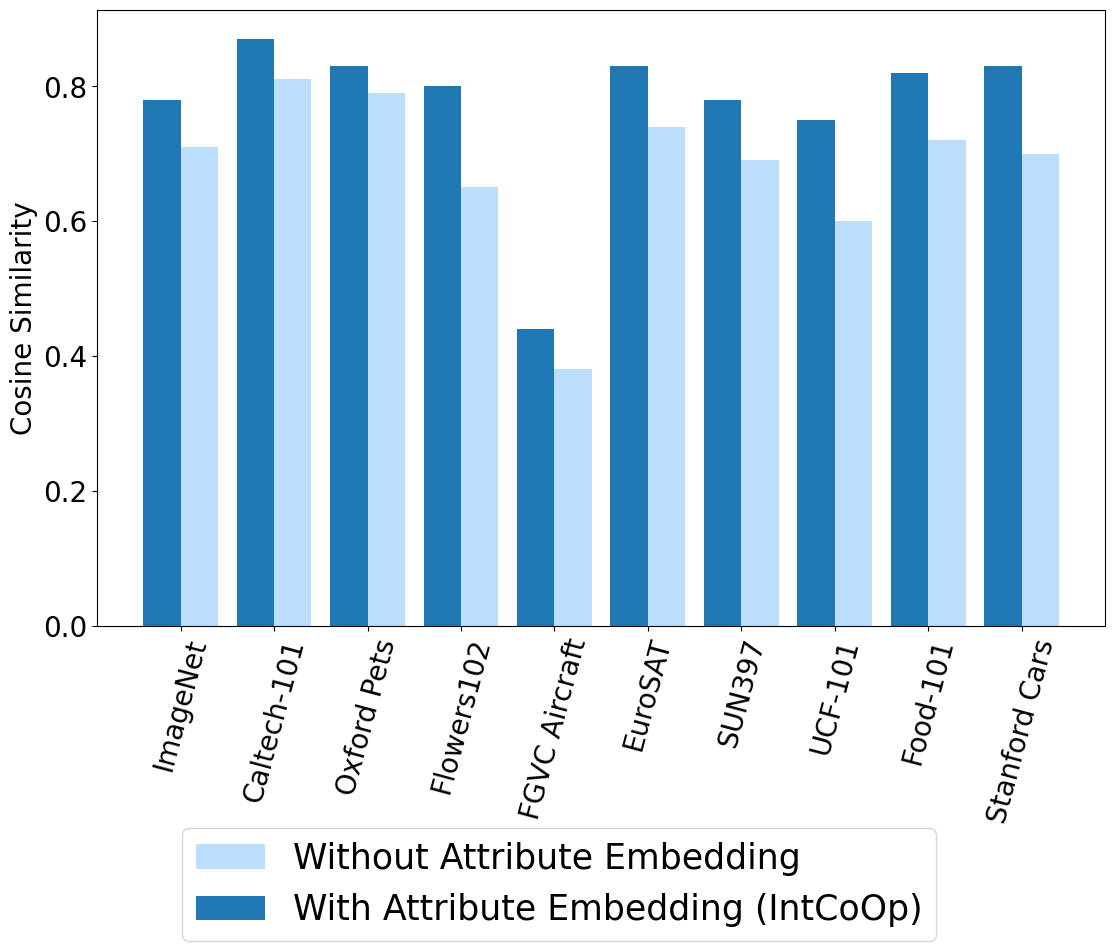}
    \caption{\small \textbf{\name generates relevant attributes during inference.} We measure the cosine similarity between the prompt embeddings with the attribute information from \name and the prompt template ``A photo of $[a]$ $[cls]$''. We find that prompt embeddings from~\name result in a higher cosine similarity with hand-crafted prompt template.}

\end{figure}

\section{Extension: Ablation on design choices}
\label{app:ablation_design}

In Table~\ref{tab:ablation_loss}, we perform an ablation study on the choice of loss functions for training \name. We find that using a $\ell_2$ loss ($f=2$) for the attribute network and a $\ell_1$ ($g=1$) regularization loss provides the best performance. Further, in Table~\ref{tab:ablation_weights}, we show ablation results for $\lambda_1$ and $\lambda_2$. Clearly setting $\lambda_1 = \lambda_2 =4$ gives the best performance.  

\section{Extension: CLIP Confidence Plots}
\begin{figure*}[t]
    \centering
    \includegraphics[width = 0.95\textwidth]{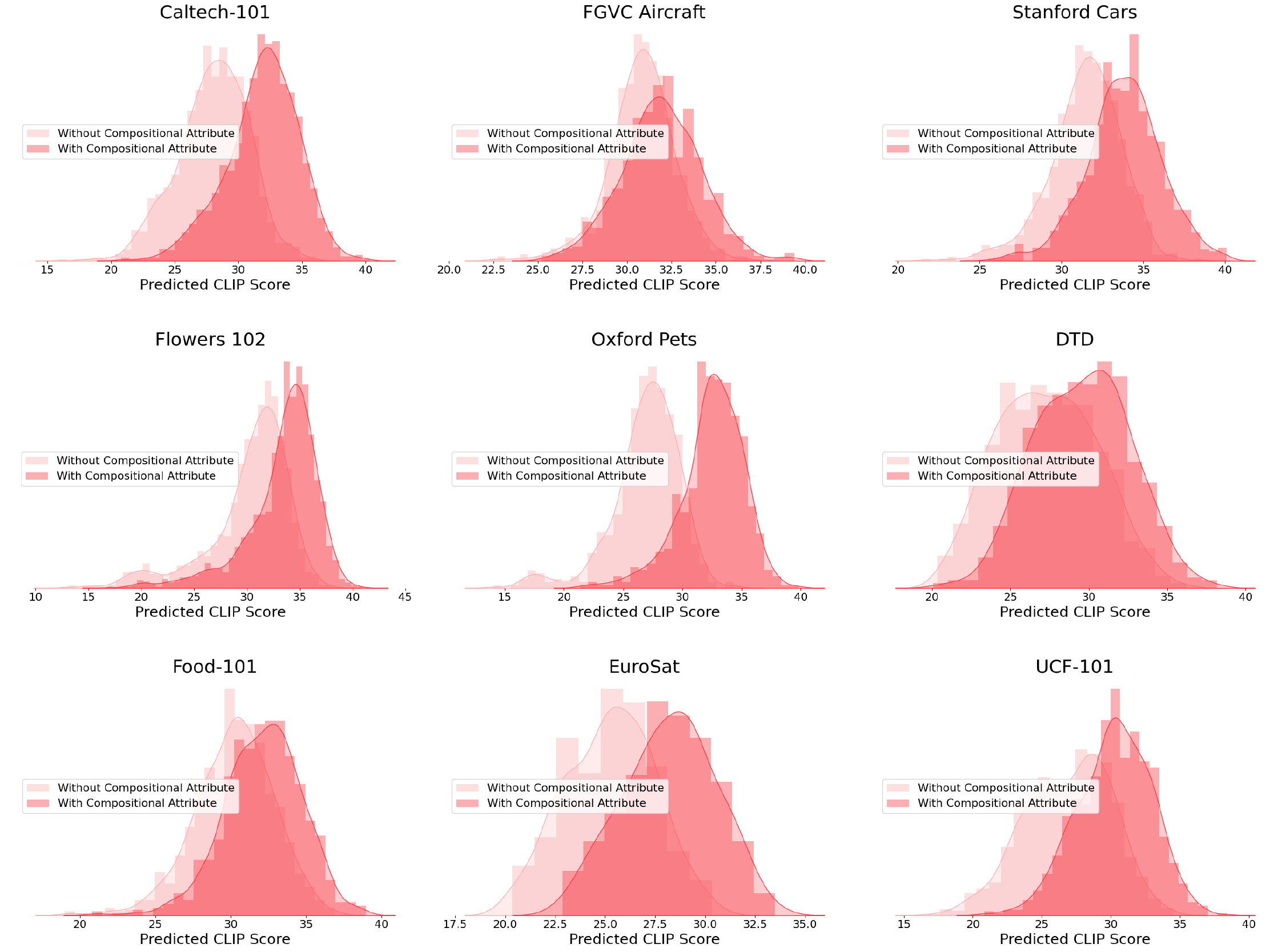}
    \caption{\textbf{CLIP Confidence plots.} Distribution plots of CLIP confidence score across different datasets highlighting the importance of incorporating compositionality information into the prompts.}
    \label{fig:clip_conf_plot}
\end{figure*}

\begin{figure*}[!h]
    \centering
    \includegraphics[width = \textwidth]{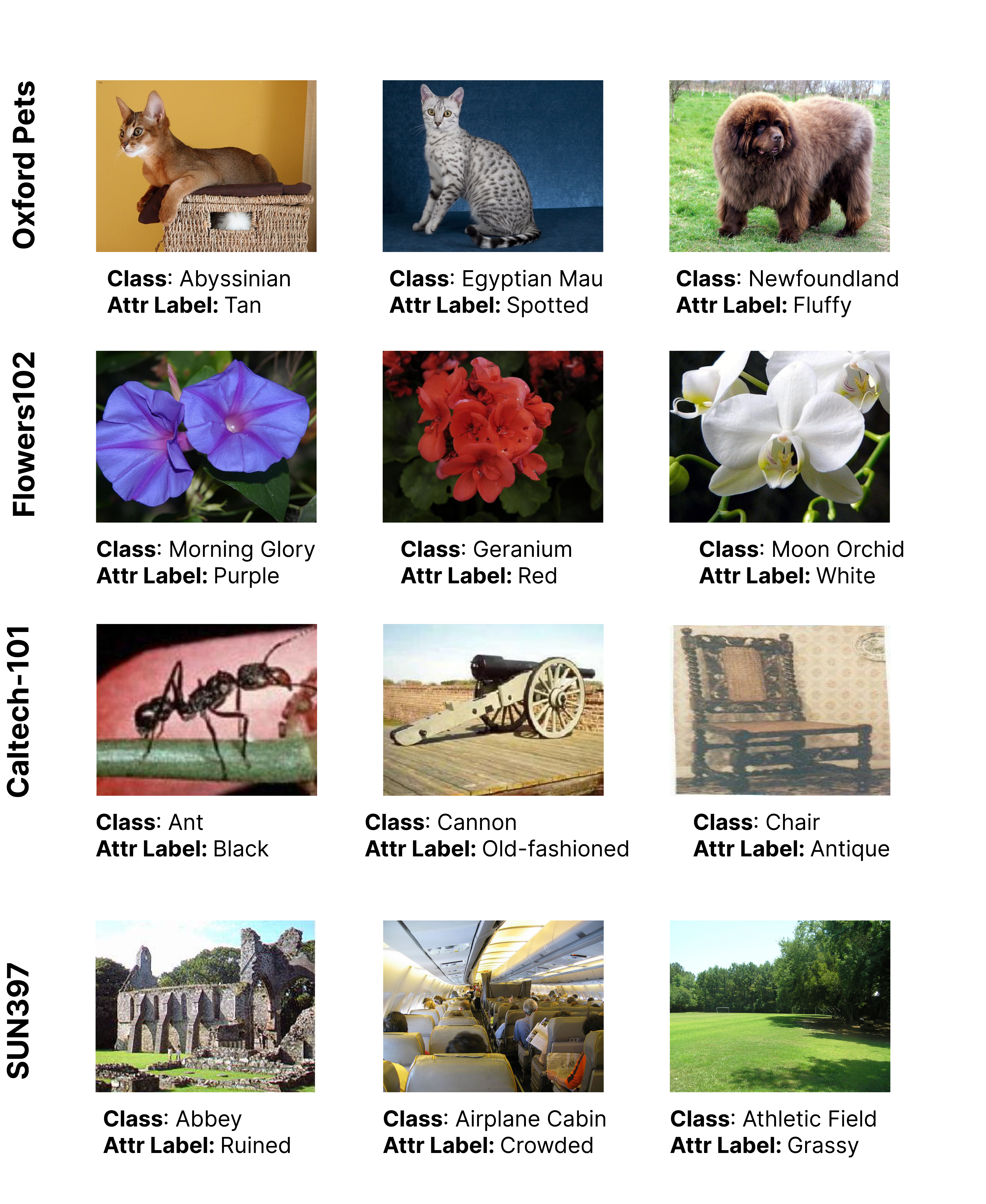}
    \caption{We visualize BLIP-2 generated attribute labels for few representative images from OxfordPets, Flowers102, Caltech-101 and SUN397 dataset.}
    \label{fig:rep_image_1}
\end{figure*}

\begin{figure*}[!h]
    \centering
    \includegraphics[width = \textwidth, keepaspectratio]{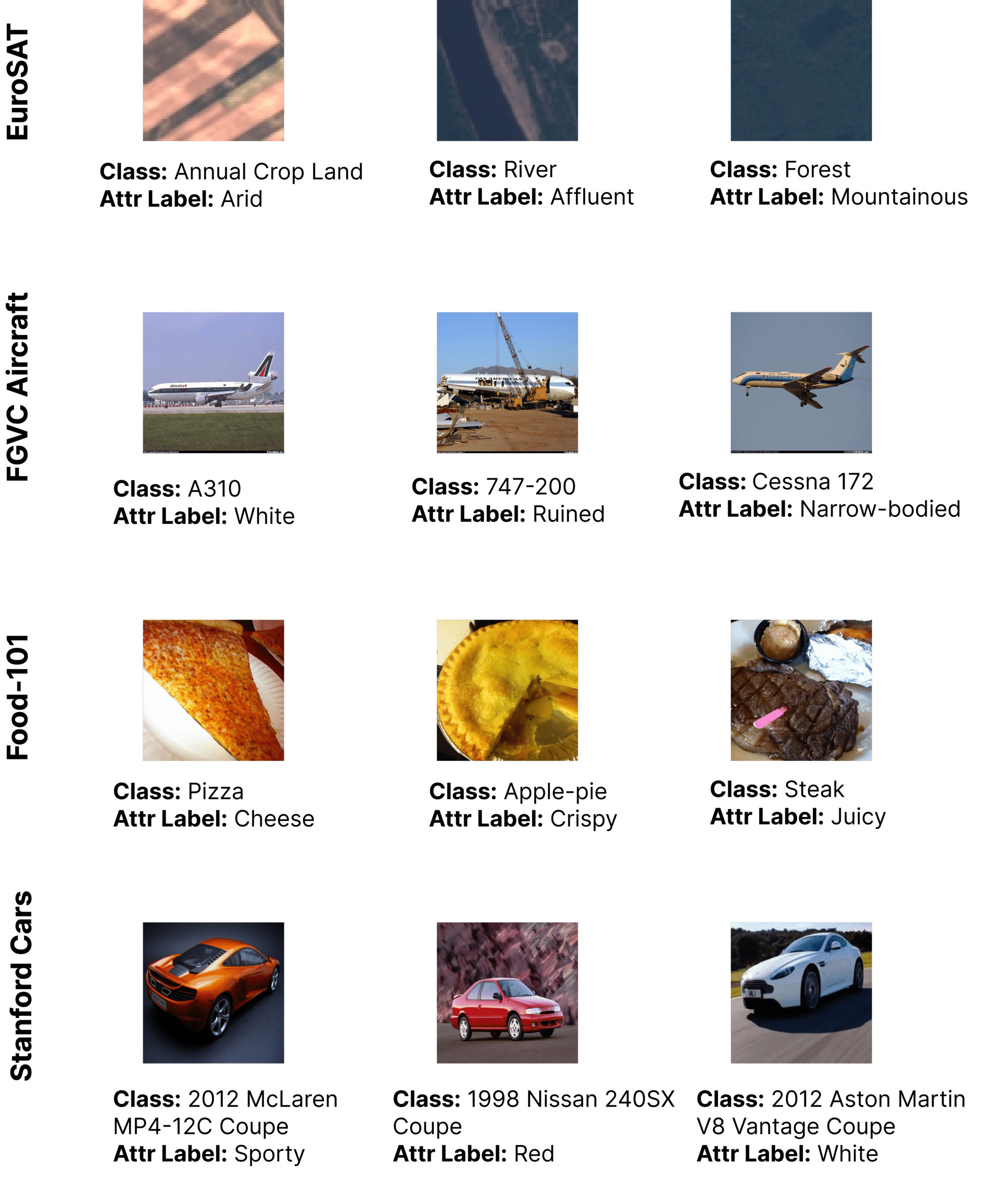}
    \caption{We visualize BLIP-2 generated attribute labels for few representative images from EuroSAT, FGVC Aircraft, Food-101 and Stanford Cars dataset.}
    \label{fig:rep_image_2}
\end{figure*}

In Figure 1 (main paper), we showed the significance of learning interpretable concepts in prompts using images from ImageNet~\cite{deng2009imagenet} dataset. Specifically, we crafted two types of prompt templates for each image: (1) one without any compositional attribute ``A photo of a $[cls]$'' and (2) other one with compositional information ``A photo of a $[a]$ $[cls]$'' where $[cls]$ represents the classname and $[a]$ denotes an attribute identified using a BLIP-2 based VQA model. Our findings indicated that the CLIP model exhibits increased confidence when prompts are enriched with compositional details of the image. Further, in Figure~\ref{fig:clip_conf_plot}, we extend this observation to additional datasets, confirming the generalizability of our results.

\newcolumntype{?}{!{\vrule width 1pt}}
\begin{table}[!t]
    \centering
    \resizebox{0.9\columnwidth}{!}{%
    \begin{tabular}{lc?c?c}
    \toprule
    & \multicolumn{1}{c}{} & \multicolumn{2}{c}{$\mathcal{L}_{\text{attr}}$}  \\
    \cmidrule{3-4}
   
    & & $g =1$ & $g =2$ \\
    \cmidrule{3-4}

    \multirow{2}{*}{$\mathcal{L}_{\text{reg}}$} & $f =1$ & 79.30/ 70.78/ 74.79 & 78.25/ 67.90/ 72.70 \\
    & $f =2$ & \cellcolor{myblue} 83.82/ 78.21/ \textbf{80.75} & 81.05/ 72.14/ 76.33 \\
    \bottomrule
    \end{tabular}%
    }
    \caption{\small\textbf{Ablation on loss functions.} We show that setting $f = 2$ and $g =1$ provides the best performance. We report the Base/ Novel/ HM accuracies for each setting. Best results based on HM performance are marked in \textbf{bold}.}
    \label{tab:ablation_loss}
\end{table}

\newcolumntype{?}{!{\vrule width 1pt}}
\begin{table}[!h]
    \centering
    \resizebox{0.8\columnwidth}{!}{%
    \begin{tabular}{l?c?c?c?c}
    \toprule
    & $\lambda_2=1$ & $\lambda_2=2$ &  $\lambda_2=4$ &  $\lambda_2=8$ \\
    \midrule
    $\lambda_1=1$ & 75.79 & 75.92 & 76.90 & 76.92\\
    \midrule
     $\lambda_1=2$ & 75.12 & 75.39 & 76.80 & 76.78\\
     \midrule
      $\lambda_1=4$ & 75.56 & 76.88 & \cellcolor{myblue}\textbf{80.75} & 77.29\\
      \midrule
       $\lambda_1=8$ & 75.97 & 76.11 & 77.31 & 77.30\\
    \bottomrule
    \end{tabular}%
    }
    \caption{Ablation results on $\lambda_1$ and $\lambda_2$. Setting $\lambda_1=4$ and $\lambda_2=4$ gives the best results. We report the HM accuracies averaged across 10 datasets for each setting. Best results based on HM performance are marked in \textbf{bold}.}
    \label{tab:ablation_weights}
\end{table}

\newcolumntype{?}{!{\vrule width 1pt}}
\newcolumntype{a}{>{\columncolor{myblue}}c}
\begin{table*}[!t]
\centering
\resizebox{\textwidth}{!}{%
\begin{tabular}{lccccccc?a}
\hline
Datasets & CoOp & CoCoOp & ProGrad & KgCoOp & MaPLe & DAPT & PLOT & \name\\
\hline
ImageNet & 69.38 & 70.55 & 70.21 & 70.19 & 70.67 & 70.80 & 70.40 & \textbf{70.81} \\
Caltech101 & 94.44 & 94.98 & 94.93 & 94.65 & 94.30 & 94.23 & 95.13 & \textbf{95.59} \\
OxfordPets & 91.30 & 93.01 & 93.21 & 93.20 & 92.05 & 92.17 & 92.55 & \textbf{96.00} \\
StanfordCars & 72.73 & 69.10 & 71.75 & 71.98 & 68.70 & 74.40 & \textbf{74.93} & \textbf{74.93} \\
Flowers102 & 91.14 & 82.56 & 89.98 & 90.69 & 80.80 & 92.37 & 91.31 & \textbf{92.54}  \\
Food101 & 82.58 & 86.64 & 85.77 & 86.59 & 86.90 & 83.60 & 86.46 & \textbf{90.60}\\
FGVCAircraft & 33.18 & 30.87 & 32.93 & 32.47 & 29.03 & 32.47 & \textbf{35.29} & 33.50 \\
SUN397 & 70.13 & 70.5 & 71.17 & 71.79 & 71.47 & 72.20 & 70.42 & \textbf{76.95} \\
EuroSAT & 68.62 & 63.83 & 70.84 & 71.06 & 54.87 & 72.73 & 80.70 & \textbf{81.21}\\
UCF101 & 77.41 & 74.99 & 77.82 & 78.40 & 73.70 & 79.40 &  \textbf{79.76} & 78.05 \\
\hline
Avg. & 75.09 & 73.69 & 75.86 & 76.10 & 72.25 & 76.38 & 77.68 & \textbf{79.01} \textbf{\textcolor{red}{(+1.34)}}\\
\hline
\end{tabular}%
}
\caption{ \textbf{\name leads to strong few-shot classification performance.} We compare \name with competitive prompt tuning approaches on a few shot learning task with $4$ samples from each class. The reported values are average performance over base and novel classes as reported by harmonic mean. We observe a $1.34\%$ improvement in average performance across 10 datasets compared to state-of-art framework PLOT~\citep{chen2023plot}. Best results are marked in \textbf{bold}.}
\label{tab:few-shot}
\end{table*}

\begin{table*}[t]
\resizebox{\textwidth}{!}{%
\begin{tabular}{llllll}
\toprule
Dataset & Classes & Train  & Val  & Test   & Description   \\
\midrule
ImageNet-1k  & 1000    & 1.28M  & N/A    & 50,000 & Contains images covering a wide range of diverse objects, scenes, and concepts.                                            \\
Caltech-101  & 101     & 4,128  & 1,649  & 2,465  & Consists of images of everyday objects commonly found in indoor and outdoor environments. \\
OxfordPets   & 37      & 2,944  & 736    & 3,669  & Comprises images of pets covering various breeds of cats and dogs in different poses.                                                                      \\
StanfordCars & 196     & 6,509  & 1,635  & 8,041  & Contains images of cars from various viewpoints, brands, and models.                                                                                       \\
Flowers102   & 102     & 4,093  & 1,633  & 2,463  & Consists of images of flowers belonging captured under varying lighting conditions and backgrounds.                           \\
Food101      & 101     & 50,500 & 20,200 & 30,300 & Consists of images depicting different types of food items from various cuisines.                                                                          \\
FGVCAircraft & 100     & 3,334  & 3,333  & 3,333  & Contains images of different airplane models captured from various viewpoints.                                                                             \\
SUN397       & 397     & 15,880 & 3,970  & 19,850 & Includes images depicting various indoor and outdoor scenes such as bedrooms, beaches, forests, and more.                                                  \\
UCF101       & 101     & 7,639  & 1,898  & 3,783  & Contains images of human actions, categorized into 101 action classes.                                                                                     \\
EuroSAT      & 10      & 13,500 & 5,400  & 8,100  & Contains satellite images capturing various land cover types including urban areas, forests, farmland, and more.  \\
\bottomrule
\end{tabular}}
\caption{Detailed description of datasets used for this study.}
\label{tab:description}
\end{table*}

\end{document}